\documentclass{article}

\usepackage[preprint,nonatbib]{files/neurips_2023}

\usepackage[utf8]{inputenc} 
\usepackage[T1]{fontenc}    
\usepackage{hyperref}       
\usepackage{url}            
\usepackage{booktabs}       
\usepackage{amsfonts}       
\usepackage{nicefrac}       
\usepackage{microtype}      
\usepackage{xcolor}         
\usepackage{natbib}
\usepackage{amsmath}
\usepackage{subfig}
\usepackage[pdftex]{graphicx}
\usepackage{float}
\usepackage{wrapfig}
\usepackage{lipsum}
\usepackage[export]{adjustbox}

\DeclareMathOperator{\x}{\mathbf{x}}

\DeclareMathOperator{\y}{\mathbf{y}}

\newcommand{\kl}[2]{D_{KL}\left(#1 \parallel #2\right)}
\DeclareMathOperator*{\E}{\mathbb{E}}

\newcommand{\tidx}[2]{#1_{#2}}
\newcommand{\didx}[2]{#1^{(#2)}}

\newcommand{\net}{\Psi}
\newcommand{\pars}{\xi}

\newcommand{\parst}[1]{\tidx{\pars}{#1}}

\newcommand{\parsdd}[1]{\didx{\pars}{#1}}
\newcommand{\alphat}[1]{\tidx{\alpha}{#1}}

\newcommand{\sender}[2]{p_{_S}\left(#1 \mid #2\right)}
\newcommand{\out}{p_{_O}}

\newcommand{\rec}{p_{_R}}
\newcommand{\inp}{p_{_I}}
\newcommand{\flow}{p_{_F}}
\newcommand{\update}{p_{_U}}

\title{Bayesian Flow Networks in Continual Learning}

\author{%
  Mateusz Pyla \\
    IDEAS~NCBR \\
    Jagiellonian~University,~Faculty~of~Mathematics~and~Computer~Science\\
    Jagiellonian~University,~Doctoral~School~of~Exact~and~Natural~Sciences\\
  \texttt{mateusz.pyla@ideas-ncbr.pl} \\
  \And
  Kamil Deja \\
  IDEAS~NCBR \\
  Warsaw~University~of~Technology
  \And
  Bartłomiej~Twardowski \\
  IDEAS~NCBR \\
  Autonomous~University of~Barcelona
  \And
  Tomasz Trzciński \\
  IDEAS~NCBR \\
  Warsaw~University~of~Technology \\ 
  Jagiellonian~University,~Faculty~of~Mathematics~and~Computer~Science
}

\begin{document}

\maketitle
\vspace{-0.5cm}
\begin{abstract}
  Bayesian Flow Networks (BFNs) has been recently proposed as one of the most promising direction to universal generative modelling, having ability to learn any of the data type. Their power comes from the expressiveness of neural networks and Bayesian inference which make them suitable in the context of continual learning. We delve into the mechanics behind BFNs and conduct the experiments to empirically verify the generative capabilities on non-stationary data.
\end{abstract}

\vspace{-0.3cm}
\section{Introduction}
\label{intro}

Diffusion models~\citep{sohl2015deep} have been progressively advancing the state-of-the-art in generative modelling, especially in the field of image processing~\citep{dhariwal2021diffusion, ho2022cascaded, kingma2021variational}. This is thanks to the usage of the diffusion processes that allows to learn complex data distributions~\citep{ho2020denoising, huang2021variational, kingma2021variational,song2019generative, song2020score}. 

However, diffusion models tend to struggle when in comes to non-continuous data. This is mostly because of the fact that denoising process for discrete, discretised or tabular data is not easy to define. 
Therefore, Alex Graves et al. recently introduced \textbf{Bayesian Flow Networks (BFNs)}~\citep{graves2023bayesian} to efficiently train the model and iteratively update the data parameters without forward pass. The general idea behind this technique is to change the way in which we model training data where instead of modelling a single instance, authors propose to output the parameters of the distribution that best fits the training data. The main motivation behind this concept is that by doing so, authors introduce an elegant way to directly model the discrete data distribution. 

In this work, we argue, that with direct modelling of parameters describing the original training data, BFNs can also be used to efficiently consolidate portions of separate data chunks. Therefore, we relate to the problem of continual learning, which tackles the ability of ML models to learn progressively as new data arrive. Bayesian update is an elegant way to manage prior belief and information from new observations. However, in Bayesian learning we often face the issue of turning theory into practical implementations, limiting the use of the Bayesian learning paradigm \citep{ding2021bridging, wang2016towards, wang2020survey}.

In this preliminary studies, we show the first benchmark of Bayesian Flow Networks in Continual Learning setup. We show how we can adapt several known techniques that prevent catastrophic forgetting in neural networks to continually train BFNs. We highlight their strengths and drawbacks and discuss future directions on how to employ BFNs to continually consolidate knowledge.

\section{Related Work}
\label{related}

\paragraph{Continual Learning (CL)} gathers various approaches in machine learning that aim at reducing catastrophic forgetting, a phenomenon where models suffer from abrupt loss in performance when retrained with additional data.
Usually there are three standard group of approaches that try to mitigate this issue: (i) architectural approaches -- methods that focus on the structure of the model itself that adds task-specific submodules to the architecture; (ii) memory approaches, methods involve storing some extra information in memory which are then used to rehearse knowledge during training in subsequent task; (iii) Regularization Approaches: that identify the important weights for the learned tasks and penalise the large updates on those weights when learning a new task~\citep{van2019three, masana2022class}.

Several of those approaches were applied in generative continual learning. In particular,~\citep{nguyen2017variational} adapt several regularisation-based methods and introduce Variational Continual Learning where additional architectural change is added with each task. In BooVAE~\citep{egorov2021boovae} an additive aggregated posterior expansion technique is used to continually trained Variational Autoencoders, while \citep{achille2018life} propose to continually disentangle data representations with VAE. Several methods train GANs in CL scenarios, e.g. using memory buffer~\citep{wu2018memory}. Most similarly to this work, in~\citep{zając2023exploring} authors benchmark the possibilities of continual-learning of diffusion models with recent CL strategies.

\begin{wrapfigure}{r}{0.53\textwidth}
\vskip  -40pt
    \begin{adjustbox}{center}
    \includegraphics[width=0.4\textwidth]{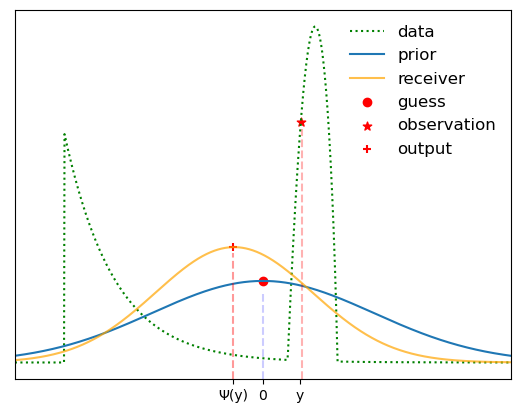}
    \end{adjustbox}
    \caption{BFNs: Our goal is to model one dimensional data distribution controlled by unknown $\xi$, which we can only sample from. We start off with some initial prior belief (0 in this case) that we are very uncertain of (blue prior). We sample an observation $y$ and add noise to obtain what we call a sender distribution -- Gaussian centered at $y$. We pass the parameters of input to the neural network obtaining output distribution, improved version enriched by jointly processes of all the variables. We comply with the noise scheduler to obtain orange receiver. We minimise the KL divergence between the sender and receiver so that our output gets closes to the samples from true data distribution.}
    \label{dynamics}
    \vskip  -37pt
\end{wrapfigure}

\section{Method}
\label{method}

Although BFNs work on different type of data, both discrete and continuous time steps, the most approachable way to understand the dynamics is through continuous data in the discrete setting and extending it as in \ref{extension}.

\subsection{Inference}
\label{inference}
We use neural network $\psi$ parameterised by $\theta$ to learn the parameters $\xi$ controlling the data distribution. The underlying distribution is complex, however we can sample from it. The general idea is to start with uneducated guess of normal distribution with high variance, and iteratively improve the estimation of data distribution with the help of the network as we sample more and more data. We assume that we model the data only with Gaussians.

As in diffusion models, we set the number of steps and we establish the accuracy scheduler managing the usefulness of noised samples for various time steps. We start from the prior belief, for instance centered at 0 with huge standard deviation. We want our network to predict better data parameters from the current estimate. To update network weights, we calculate the gradient as a KL divergence between the predicted and original data in their noised forms. 
While explaining the mathematical formulation behind the process, we point out concrete example in~\ref{dynamics}.

More rigorously, we threat each data variable separately. Due to the independence, the input distribution can be expressed as the product of one dimensional distributions.
\begin{alignat}{1}
\inp(\x \mid \pars) = \Pi_{d=1}^D \inp(\didx{x}{d} \mid \parsdd{d})\label{input}
\end{alignat}
Rather than data points, we receive the noisy samples forming the normal distribution centered at the true values with variance purely depended on the accuracy scheduler.
\begin{alignat}{1}
\sender{\y}{\x;\alpha} = \Pi_{d=1}^D \sender{\didx{y}{d}}{\didx{x}{d}; \alpha}\label{sender}
\end{alignat}
Since we proceed each dimension independently, we need global feedback coming from the interactions between variables. The role of neural network is to update the guess knowing the previous belief, so that we can better decode the sended sample.
\begin{alignat}{1}
\out(\x \mid \pars, t) = \Pi_{d=1}^D \out(\didx{x}{d} \mid \didx{\net}{d}(\pars, t))\label{output}
\end{alignat}
Since we are not aware of the true $\didx{x}{d}$, knowing $\sender{\cdot}{\didx{x}{d}; \alpha}$ we can only marginalise over all possible values $\didx{x'}{d}$ weighted by the output probability, obtaining the receiver: 
\begin{alignat}{1}
\rec(\y \mid \pars; t, \alpha) &= \E_{\out(\x' \mid \pars; t)}\sender{\y}{\x'; \alpha}\label{receiver}
\end{alignat}
Iteratively, for the next time steps, we apply Bayesian updates to improve the input distribution (that accumulates only local knowledge about a single dimension) by the acquired knowledge from receiver (that encodes global knowledge about interactions between dimensions). Both input and sender are Gaussian distributions and factorised independently, hence the update is straightforward: $\rho_{i+1} = \rho_i + \alpha$ and $\mu_{i+1} = \frac{\rho_i \mu_i+\alpha y}{\rho_{i+1}}$
Once we set the number of steps to infinity, under mild conditions for the scheduler, we are able to efficiently compute the dynamics:
\begin{alignat}{1}
\flow(\pars \mid \x ; t) = \update(\pars \mid \parst{0}, \x; \beta(t))\label{param_flow_dist}.
\end{alignat}

There is a freedom in choosing the underlying network, as long as it returns the new parameters of data (U-Net \citep{ronneberger2015unet}, Transformers \citep{vaswani2023attention}, TabTransformer \citep{huang2020tabtransformer}) and inference conditions on the time step.

The proposed scheduler is of form $\beta(t) \doteq \sigma_1^{-2t}-1$ for $t\in[0,1]$ yielding accuracy rate $\alpha(t)=\frac{2\log\sigma_1}{\sigma_1^{2t}}$. When $\alpha$ is 0, the model is uninformative of samples and confidence increases with the higher values. $\sigma_1$ is the hyperparameter standing for standard deviation at the final time.

\vspace{-0.3cm}
\subsection{Training}
\label{training}

Loss function can be intuitively understood as costs of revealing the underlying data distribution with the least possible effort or information. Our objective is to match output distribution to the data distribution indirectly by optimizing KL divergence between their noisy versions. Specifically, we minimise KL divergence between sender and receiver distributions:
$\kl{p_{_S}}{\rec}$ 
in order to bring output predictions closer and closer to the true data values.
\begin{align}
L^n(\x) \doteq \E_{p(\parst{1},\dots,\parst{n-1})}\sum_{i=1}^n \kl{\sender{\cdot}{\x ; \alphat{i}}}{\rec(\cdot \mid \parst{i-1} ; t_{i-1}, \alphat{i})} \label{disc_t_loss_n_step},
\end{align}
This loss indirectly optimises our true goal:
\begin{align}
L^r(\x) = -\E_{\flow(\pars \mid \x, 1)} \ln \out(\x \mid \pars; 1).
\end{align}
Let us note that this kind of form follows information-theory interpretation: we minimise the number of nats required to transmits a sample between two distributions. 

\vspace{-0.3cm}
\subsection{BFN in Continual Learning}
\label{bfncl}

We propose to extend the basic idea of BFNs in order to benchmark it with several known continual-learning strategies. In particular, we start with a simple regularisation strategy, where we prevent model in subsequent tasks to diverge from the previous by penalising $\mathcal{L}_1$ or $\mathcal{L}_2$ norm.

We compare the regularisation approach with two rehearsal-based methods. In the first one we employ a simple buffer-based rehearsal where we store a subset of previous data examples in a buffer and use them together with new data samples when retraining a model on new tasks. In the second one, taking advantage of the generative model we continually train, we propose to generate examples from previous tasks in order to use them as rehearsal samples in a generative replay approach.

\vspace{-0.3cm}
\section{Experiments}
\label{experiments}

We evaluate the performance of BFNs in Continual Learning using the standard MNIST dataset and our new scenario with tabular data on US flights connections in 2013 \citep{flights2013data}. One of the most common setting in which we are able to assess the continual learning capabilities of the proposed model is to split the training dataset into disjoint chunks and perform the training in a sequential way. In Class-Incremental Learning set up, each task often contains the same . Each task $\tau_i$ is associated with a dataset $\mathcal{D}_i$, and the objective is to model the distribution of $\mathcal{D}_i$.

In particular, we split the MNIST in CIL setting $5 \times 2$ by dividing it into 5 tasks, each binary classification of two consecutive digits. Following~\citep{graves2023bayesian}, we also binarise the images. In flight dataset we divide group flights by month of the journey obtaining 12 tasks. 

\subsection{Image dataset}
\begin{figure}[h]
  \centering
  \begin{minipage}{0.31\linewidth}
      \begin{figure}[H]
          \includegraphics[width=\linewidth]{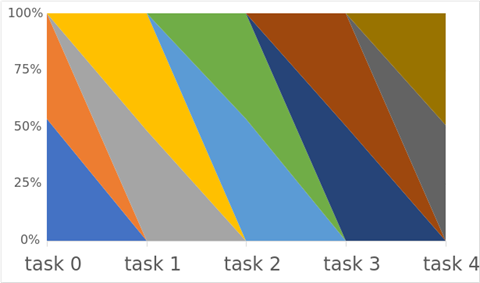}
      \end{figure}
  \end{minipage}
  \begin{minipage}{0.31\linewidth}
      \begin{figure}[H]
          \includegraphics[width=\linewidth]{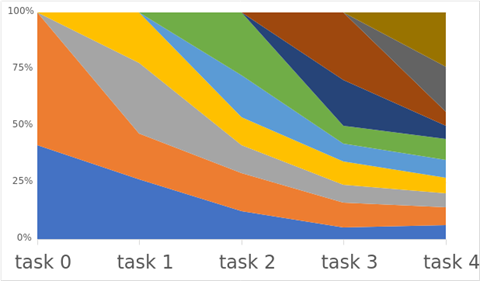}
      \end{figure}
  \end{minipage}
  \begin{minipage}{0.31\linewidth}
      \begin{figure}[H]
          \includegraphics[width=\linewidth]{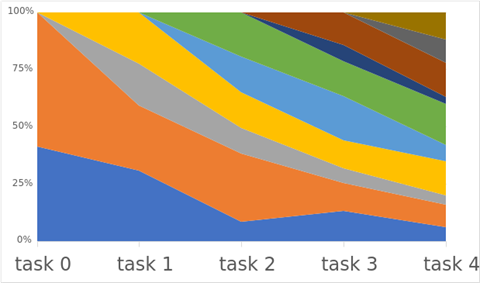}
      \end{figure}
  \end{minipage}
  \begin{minipage}{0.04\linewidth}
      \begin{figure}[H]
          \includegraphics[scale=0.2]{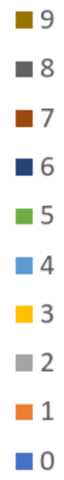}
      \end{figure}
  \end{minipage}
  \caption{\label{fig:results_mnist}Results of classification of the generated images (a) left: using finetuning (b) middle: memory-based (c) right: generative-based method. (d) The colours corresponds to percentage share of consecutive digits across the sequential training.}
\end{figure}

\begin{figure}[h]
  \centering
  \begin{minipage}{0.32\linewidth}
      \begin{figure}[H]
          \includegraphics[width=\linewidth]{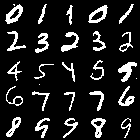}
      \end{figure}
  \end{minipage}
  \begin{minipage}{0.32\linewidth}
      \begin{figure}[H]
          \includegraphics[width=\linewidth]{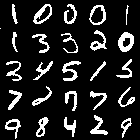}
      \end{figure}
  \end{minipage}
  \begin{minipage}{0.32\linewidth}
      \begin{figure}[H]
          \includegraphics[width=\linewidth]{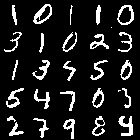}
      \end{figure}
  \end{minipage}
  \caption{\label{fig:results_mnist_gen}Results of image generations for (a) finetuning; (b) memory-based approach; (c) generative-based approach. Each row corresponds to the consecutive task.}
\end{figure}

In Figure~\ref{fig:results_mnist}, we present the results of our experiments with MNIST dataset. To measure the catastrophic forgetting, after each task, we generate 1000 examples and report the share of each class as measured by the externally trained classifier. As visible, in finetuning (without any CL strategy), we can observe catastrophic forgetting as with each new task model abruptly forgets how to generate examples from the previous classes. On the other hand, both: buffer-based and generative-based replay prevent catastrophic forgetting, as even after the last task, we can observe some generations of classes from the first task. For qualitative analysis we provide some samples in Figure~\ref{fig:results_mnist_gen}.

\vspace{-0.4cm}

\vspace{-0.4cm}
\section{Tabular data}
\begin{figure}[h]
  \centering
  \begin{minipage}{0.45\linewidth}
      \begin{figure}[H]
          \includegraphics[width=0.8\linewidth]{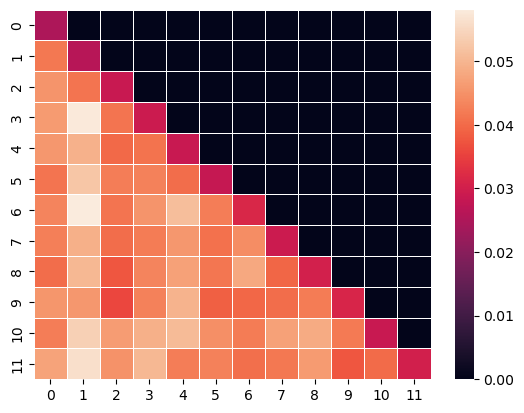}
      \end{figure}
  \end{minipage}
  \begin{minipage}{0.45\linewidth}
      \begin{figure}[H]
          \includegraphics[width=0.8\linewidth]{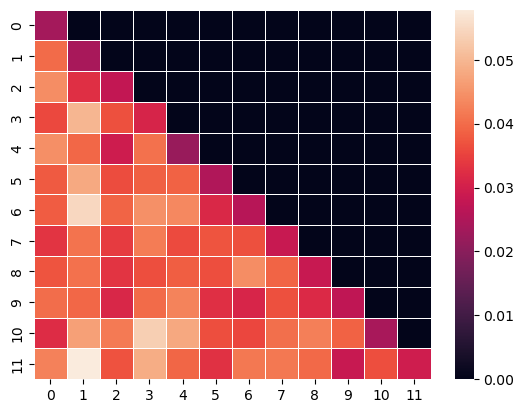}
      \end{figure}
  \end{minipage}
  \caption{\label{fig:results_tab}Results of loss applied on test images (a) left: using finetuning (b) right: generative-replay method. Loss induced by BFNs measures bits by dimension and offers (negative log-) likelihood estimation interpretation. On $Y$ axis incrementally we are proceed with tasks, whereas values on $X$ axis corresponds to data batches from indicated tasks.}
\end{figure}

To evaluate the performance of BFNs in modelling categorical data in the continual learning scenario, we refer to the problem of tabular data modelling. The results are presented in \autoref{fig:results_tab}. We inspect the test loss metric as a proxy for model surprise of provided data.
\vspace{-0.2cm}

\vspace{-0.5cm}
\section{Conclusion, Limitations and Future work}
\label{future}
Bayesian Flow Networks (BFNs) are exciting family of generative models that are able to deal with various type of data. In this work we highlight that modelling data distribution parameters does not prevent those models from catastrophic forgetting. However, BFNs can benefit from known CL strategies such as rehearsal and generative replay. In our future works, we plan to explore further how we can benefit from the sweet combination of Bayesian update, with neural modelling in order to continually adjust parameters of the data distrubution.

\bibliographystyle{plain}

\begin{thebibliography}{99}

\bibitem[Alessandro Achille, Tom Eccles, Loic Matthey, Chris Burgess, Nicholas Watters, Alexander Lerchner, and Irina Higgins.]{achille2018life} Alessandro Achille, Tom Eccles, Loic Matthey, Chris Burgess, Nicholas Watters, Alexander Lerchner, and Irina Higgins. Life-long disentangled representation learning with cross-domain latent homologies. {\em Advances in Neural Information Processing Systems}, 31, 2018.

\bibitem[Prafulla Dhariwal, Alexander Nichol.]{dhariwal2021diffusion}Prafulla Dhariwal and Alexander Nichol. Diffusion models beat {GANs} on image synthesis.{\em Advances in Neural Information Processing Systems}, 34, 2021.

\bibitem[Nan Ding, Xi~Chen, Tomer Levinboim, Sebastian Goodman, and Radu Soricut.]{ding2021bridging}Nan Ding, Xi~Chen, Tomer Levinboim, Sebastian Goodman, and Radu Soricut. Bridging the gap between practice and pac-bayes theory in few-shot meta-learning, 2021.

\bibitem[CC0:Public Domain, 2013]{flights2013data} CC0:~Public Domain. Flights dataset in 2013. \url{https://www.kaggle.com/datasets/mahoora00135/flights/data}, 2013.

\bibitem[Evgenii Egorov, Anna Kuzina, and Evgeny Burnaev.] {egorov2021boovae} Evgenii Egorov, Anna Kuzina, and Evgeny Burnaev. Boovae: Boosting approach for continual learning of vae. {\em Advances in Neural Information Processing Systems}, 34, 2021.

\bibitem[Alex Graves, Rupesh~Kumar Srivastava, Timothy Atkinson, and Faustino Gomez.]{graves2023bayesian} Alex Graves, Rupesh~Kumar Srivastava, Timothy Atkinson, and Faustino Gomez. Bayesian flow networks, 2023.

\bibitem[Jonathan Ho, Ajay Jain, and Pieter Abbeel.]{ho2020denoising} Jonathan Ho, Ajay Jain, and Pieter Abbeel. Denoising diffusion probabilistic models. {\em Advances in Neural Information Processing Systems}, 33:6840--6851, 2020.

\bibitem[Jonathan Ho, Chitwan Saharia, William Chan, David~J Fleet, Mohammad Norouzi, and Tim Salimans.]{ho2022cascaded} Jonathan Ho, Chitwan Saharia, William Chan, David~J Fleet, Mohammad Norouzi, and Tim Salimans. Cascaded diffusion models for high fidelity image generation.{\em Journal of Machine Learning Research}, 23(47):1--33, 2022.

\bibitem[Chin-Wei Huang, Jae~Hyun Lim, and Aaron~C Courville.]{huang2021variational} Chin-Wei Huang, Jae~Hyun Lim, and Aaron~C Courville. A variational perspective on diffusion-based generative models and score matching. {\em Advances in Neural Information Processing Systems}, 34, 2021.

\bibitem[Xin Huang, Ashish Khetan, Milan Cvitkovic, and Zohar Karnin.]{huang2020tabtransformer}
Xin Huang, Ashish Khetan, Milan Cvitkovic, and Zohar Karnin.
\newblock Tabtransformer: Tabular data modeling using contextual embeddings, 2020.

\bibitem[Diederik Kingma, Tim Salimans, Ben Poole, and Jonathan Ho.
\newblock Variational diffusion models.]{kingma2021variational}
Diederik Kingma, Tim Salimans, Ben Poole, and Jonathan Ho. Variational diffusion models. In M.~Ranzato, A.~Beygelzimer, Y.~Dauphin, P.S. Liang, and J.~Wortman Vaughan, editors, {\em Advances in Neural Information Processing Systems}, volume~34, pages 21696--21707. Curran Associates, Inc., 2021.

\bibitem[Marc Masana, Xialei Liu, Bart{\l}omiej Twardowski, Mikel Menta, Andrew~D Bagdanov, and Joost van~de Weijer.]{masana2022class}
Marc Masana, Xialei Liu, Bart{\l}omiej Twardowski, Mikel Menta, Andrew~D Bagdanov, and Joost van~de Weijer. Class-incremental learning: survey and performance evaluation on image classification. {\em IEEE Transactions on Pattern Analysis and Machine Intelligence}, 2022.

\bibitem[Cuong~V Nguyen, Yingzhen Li, Thang~D Bui, and Richard~E Turner.]{nguyen2017variational}
Cuong~V Nguyen, Yingzhen Li, Thang~D Bui, and Richard~E Turner.
\newblock Variational continual learning.
\newblock In {\em International Conference on Learning Representations}, 2018.

\bibitem[Olaf Ronneberger, Philipp Fischer, and Thomas Brox.]{ronneberger2015unet}
Olaf Ronneberger, Philipp Fischer, and Thomas Brox.
\newblock U-net: Convolutional networks for biomedical image segmentation, 2015.

\bibitem
[Jascha Sohl-Dickstein, Eric Weiss, Niru Maheswaranathan, and Surya Ganguli.]{sohl2015deep} Jascha Sohl-Dickstein, Eric Weiss, Niru Maheswaranathan, and Surya Ganguli. Deep unsupervised learning using nonequilibrium thermodynamics. In {\em International Conference on Machine Learning}, pages 2256--2265. PMLR, 2015.

\bibitem[Yang Song, Stefano Ermon.]{song2019generative}
Yang Song and Stefano Ermon.
\newblock Generative modeling by estimating gradients of the data distribution.
\newblock {\em Advances in Neural Information Processing Systems}, 32, 2019.

\bibitem[Yang Song, Jascha Sohl-Dickstein, Diederik~P Kingma, Abhishek Kumar, Stefano Ermon, and Ben Poole.]{song2020score}
Yang Song, Jascha Sohl-Dickstein, Diederik~P Kingma, Abhishek Kumar, Stefano Ermon, and Ben Poole.
\newblock Score-based generative modeling through stochastic differential equations.
\newblock In {\em International Conference on Learning Representations}, 2020.

\bibitem[Gido~M Van~de Ven, Andreas~S Tolias.]{van2019three}
Gido~M Van~de Ven and Andreas~S Tolias.
\newblock Three scenarios for continual learning.
\newblock {\em arXiv preprint arXiv:1904.07734}, 2019.

\bibitem[Ashish Vaswani, Noam Shazeer, Niki Parmar, Jakob Uszkoreit, Llion Jones, Aidan~N. Gomez, Lukasz Kaiser, and Illia Polosukhin.]{vaswani2023attention}
Ashish Vaswani, Noam Shazeer, Niki Parmar, Jakob Uszkoreit, Llion Jones, Aidan~N. Gomez, Lukasz Kaiser, and Illia Polosukhin.
\newblock Attention is all you need, 2023.

\bibitem[Hao Wang, Dit-Yan Yeung.]{wang2016towards}
Hao Wang and Dit-Yan Yeung.
\newblock Towards bayesian deep learning: A framework and some existing methods.
\newblock {\em IEEE Transactions on Knowledge and Data Engineering}, 28(12):3395--3408, 2016.

\bibitem[Hao Wang, Dit-Yan Yeung.]{wang2020survey}
Hao Wang and Dit-Yan Yeung.
\newblock A survey on bayesian deep learning.
\newblock {\em ACM computing surveys (csur)}, 53(5):1--37, 2020.

\bibitem[Chenshen Wu, Luis Herranz, Xialei Liu, Yaxing Wang, Joost van~de Weijer, and Bogdan Raducanu.]{wu2018memory}
Chenshen Wu, Luis Herranz, Xialei Liu, Yaxing Wang, Joost van~de Weijer, and Bogdan Raducanu.
\newblock Memory replay gans: Learning to generate new categories without forgetting.
\newblock In {\em NeurIPS}, 2018.

\bibitem[Michał Zając, Kamil Deja, Anna Kuzina, Jakub~M. Tomczak, Tomasz Trzciński, Florian Shkurti, and Piotr Miłoś.]{zając2023exploring}
Michał Zając, Kamil Deja, Anna Kuzina, Jakub~M. Tomczak, Tomasz Trzciński, Florian Shkurti, and Piotr Miłoś.
\newblock Exploring continual learning of diffusion models, 2023.

\end{thebibliography}

\section*{Appendix}

\subsection{Extension}
\label{extension}
For discrete or discretised case, we assume that we model the distributions by $\alpha$ parameters, which is the vector of length equal to the number of possible values, i.e. $\alpha_i$ is the probability of generating $i^{th}$ value. For more technicalities, we refer to chapter 5 in the original paper.

In order to obtain closed-form update formulas, under mind conditions we set the accuracy schedule $ \beta $ which is just compound sum of accuracy rates, i.e. $ \beta(t) \doteq \int_{t'=0}^t \alpha_{t'}dt'$ as in diffusion models. Then, through the series of mathematical derivations, we are able to calculate the overall Bayesian Flow update given the time step as in equation 205 in the original paper.

\subsection{Additional results - image generations}
\vspace{-0.2cm}

\subsubsection{Additional results -- regularisation-based approach.}

We also notice the quality degradation for regularisation-based approaches. We observed two cases, either the $\lambda$ coefficient corresponding to the weight of regularisation component is too small and therefore not significant, or there is quality degradation for image generation.

\begin{figure}[H]
  \includegraphics[width=\linewidth]{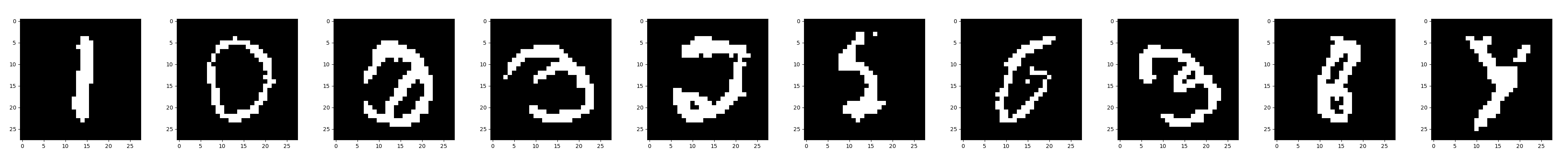}
  \caption{\label{fig:results_mnist_reg} Results of image generation for regularisation-based approach. For each task, two samples were generated. }
\end{figure}

\paragraph{Experiment details}
\label{details}

For work on MNIST, we apply standard U-Net, widely applied for diffusion models with 64 model channels, 2 resnet blocks and attention resolution 32,16,8. The size of the model is 6.1M parameters.

For work on tabular data (5 categorical, 9 numerical columns), we use TabTransformer with dimension 32, depth 6 and 8 heads. Overall models has around 600k learnable parameters. There are roughly 320k data points.

\end{document}